\documentclass[runningheads]{llncs}

\usepackage{eccv}

\usepackage{eccvabbrv}

\usepackage{graphicx}
\usepackage{booktabs}
\usepackage{tabularx}
\usepackage[accsupp]{axessibility}  

\usepackage{graphicx,verbatim}
\usepackage{algorithm}
\usepackage{algpseudocode}   
\usepackage{amsmath,amssymb} 
\usepackage{booktabs}
\usepackage{multirow}

\newcommand{\best}[1]{\textbf{#1}}
\newcommand{\bestfig}[1]{\colorbox{green!15}{\textbf{#1}}}

\usepackage{tcolorbox}
\usepackage{amssymb}
\usepackage{enumitem}
\definecolor{resBlue}{HTML}{3A7CC3}      
\definecolor{actAmber}{HTML}{D4960A}     
\definecolor{refRed}{HTML}{C0392B}       
\definecolor{skillGreen}{HTML}{2E7D32}   
\definecolor{memGold}{HTML}{B8860B}      
\definecolor{toolPurple}{HTML}{7B5EA7}   
\definecolor{passGreen}{HTML}{27AE60}    
\definecolor{failRed}{HTML}{C0392B}      

\tcbset{
    casebox/.style={
        colback=lightgray, colframe=#1, 
        fonttitle=\bfseries\small, 
        boxrule=0.8pt, arc=2pt,
        left=4pt, right=4pt, top=2pt, bottom=2pt,
        fontupper=\scriptsize
    }
}

\usepackage{tikz}
\usetikzlibrary{positioning, fit, backgrounds, calc, arrows.meta, decorations.pathreplacing}

\definecolor{propbg}{HTML}{EBF5FB}
\definecolor{propborder}{HTML}{2980B9}
\definecolor{rankbg}{HTML}{E8F8F5}
\definecolor{rankborder}{HTML}{1ABC9C}
\definecolor{parambg}{HTML}{FEF9E7}
\definecolor{paramborder}{HTML}{F39C12}
\definecolor{headerbg}{HTML}{2C3E50}
\definecolor{darktext}{HTML}{2C3E50}

\usepackage{hyperref}

\usepackage{orcidlink}

\begin{document}

\title{TopoAgent: An Agentic Framework for Automated Topology Learning in Medical Imaging}

\titlerunning{TopoAgent for Automated Topology Learning}
\authorrunning{G. Meng et al.}

\newcommand{\corr}{\textsuperscript{\ensuremath{\dagger}}}

\author{
Guangyu Meng\inst{1} \and
Pengfei Gu\inst{2}\corr \and
Xueyang Li\inst{1} \and
Yiyu Shi\inst{1} \and
Erin Wolf Chambers\inst{1}\corr \and
Danny Z. Chen\inst{1}\corr
}

\institute{ Dept.\ of Computer Science and Engineering, University of Notre Dame, Notre Dame, IN, USA
\and Dept.\ of Computer Science, The University of Texas Rio Grande Valley, Edinburg, TX, USA\\[0.3em]
{\tt\small \{gmeng, xli34, Yiyu.Shi.31, echambe2, dchen\}@nd.edu, \quad pengfei.gu01@utrgv.edu}\\[0.3em]
}

\maketitle

\begingroup
\renewcommand{\thefootnote}{\ensuremath{\dagger}}
\begin{NoHyper}
\footnotetext{\scriptsize Corresponding authors.}
\end{NoHyper}
\endgroup

\begin{abstract}
Topological data analysis (TDA), particularly persistent homology (PH), captures geometric structural properties in medical images (e.g., connected components, loops, shape characteristics), which conventional pixel-level deep learning approaches often neglect. While many topological descriptors are known for converting persistence diagrams (PDs) or raw images into topological feature vectors, existing methods mostly default to a single fixed descriptor (e.g., persistence images), leaving the diversity of topological representations largely unexplored. To the best of our knowledge, there is no known large language model (LLM)-based agentic framework that can automatically determine the most suitable topological descriptors for a given image dataset and produce the corresponding topological feature vectors for downstream tasks. To fill this gap, we propose \textbf{TopoAgent}, an LLM-based agentic framework that automates topology learning for medical image analysis. 
TopoAgent operates through a Perception--Reasoning--Action--Reflection loop supported by 21 domain-specific tools and dual memory that accumulates experience across runs. Its skill set is distilled from systematic evaluation of 15 topological descriptors across 26 datasets with six classifiers. TopoAgent analyzes input images and their topological characteristics, reasons about which topological descriptors best suit the input, and determines the optimal descriptor and its configuration, all without task-specific training. To evaluate TopoAgent, we introduce \textbf{TopoBenchmark}, a frozen benchmark of 113,182 samples from 26 medical image datasets spanning five object types: 
cells, glands and lumens, organ shapes, vessel trees, and surface lesions. Experiments show that TopoAgent obtains 68.21\% average balanced accuracy, outperforming the strongest baseline by 9.32\% and general-purpose LLMs equipped with the same tools by over 21\%.
\end{abstract}

\section{Introduction}
\label{sec:intro}
\begin{figure*}[!t]
\centering
\setlength{\abovecaptionskip}{4pt}
\setlength{\belowcaptionskip}{0pt}

\resizebox{1\textwidth}{!}{%
    \begin{minipage}{\textwidth}
\tcbset{
  rbox/.style={
    fontupper=\normalsize, left=3pt, right=3pt, top=3pt, bottom=3pt,
    before upper=\raggedright, boxrule=0.6pt, valign=top,
    equal height group=topoAgentRow
  }
}

\textbf{(a) Accuracy of different topological descriptors on five datasets}\par\vspace{4pt}
\noindent\resizebox{\linewidth}{!}{%
\renewcommand{\arraystretch}{1.0}%
\setlength{\tabcolsep}{5pt}%
\begin{tabular}{@{}c@{\hspace{6pt}}l cccccc c@{}}
    \toprule
    & \textbf{Dataset} & \textbf{PS~\cite{ali2023survey}} & \textbf{PI~\cite{adams2017persistence}} & \textbf{TF~\cite{perea2023approximating}} & \textbf{ATOL~\cite{royer2021atol}} & \textbf{MK~\cite{mecke2000additivity}} & \textbf{LBP~\cite{ojala2002multiresolution}} & \textbf{Best} \\
    \midrule
    \raisebox{-0.35em}{\includegraphics[height=0.58cm]{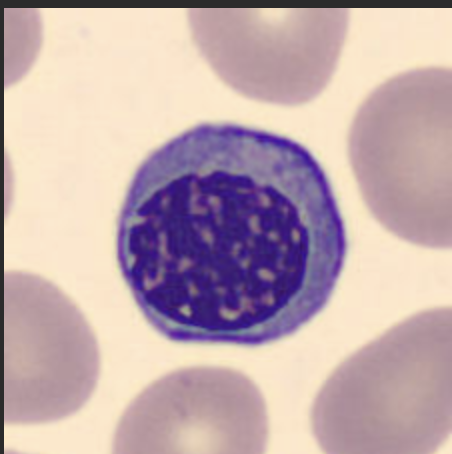}}
    & BloodMNIST~\cite{medmnistv2}   & 96.75 & 90.63 & \bestfig{97.92} & 97.64 & 97.75 & 90.03 & TF \\
    \raisebox{-0.35em}{\includegraphics[height=0.58cm]{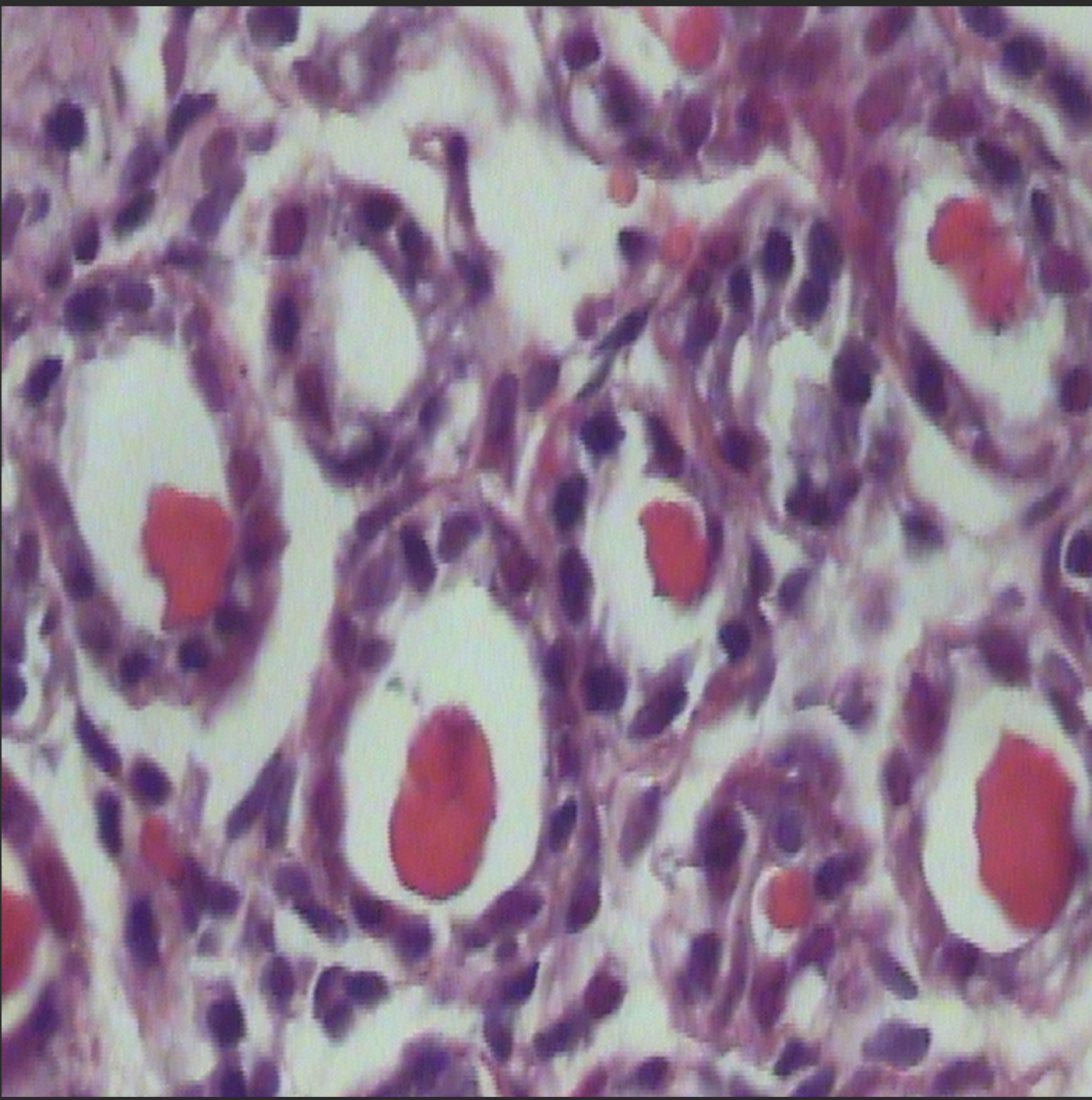}}
    & BreakHis~\cite{breakhis}     & \bestfig{88.76} & 67.92 & 79.64 & 87.03 & 83.91 & 77.02 & PS \\
    \raisebox{-0.35em}{\includegraphics[height=0.58cm]{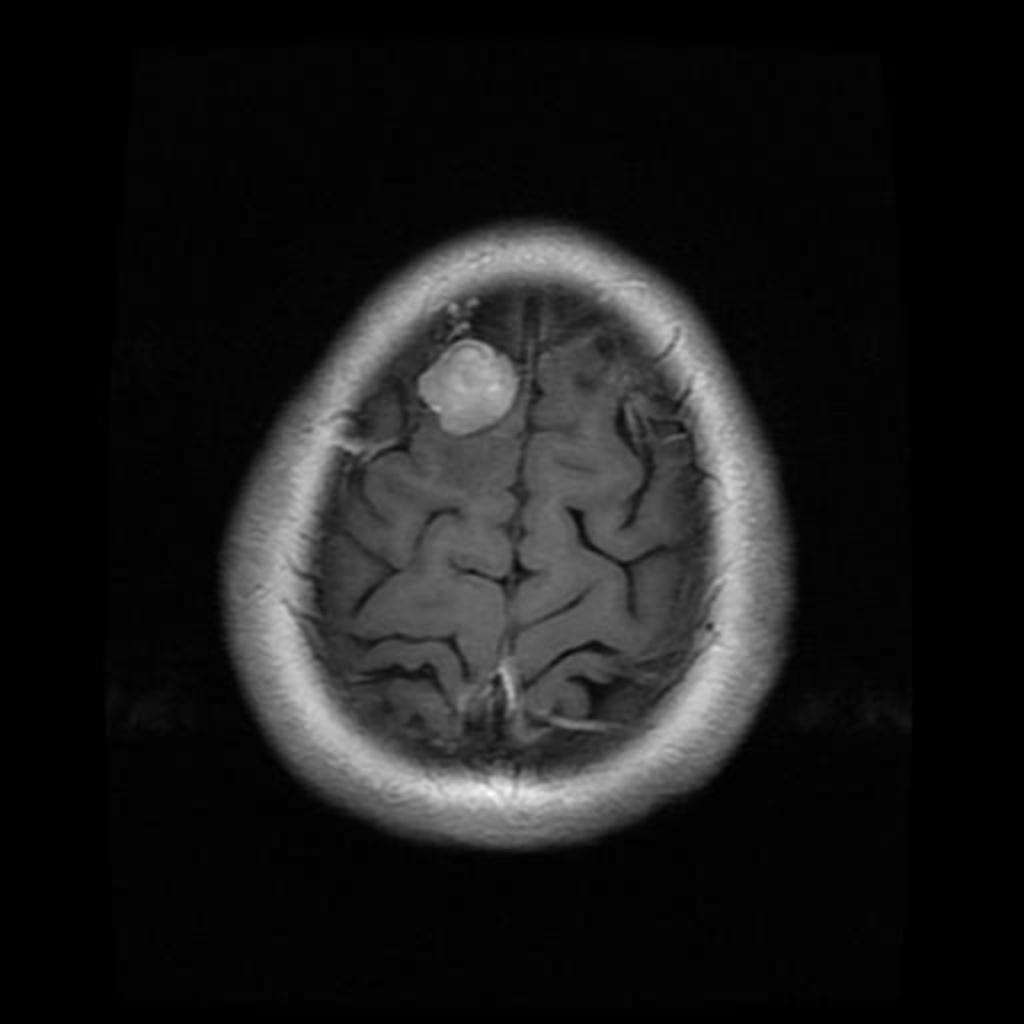}}
    & BrainTumor~\cite{braintumormri}   & 95.10 & 83.49 & 95.12 & \bestfig{96.16} & 94.71 & 94.98 & ATOL \\
    \raisebox{-0.35em}{\includegraphics[height=0.58cm]{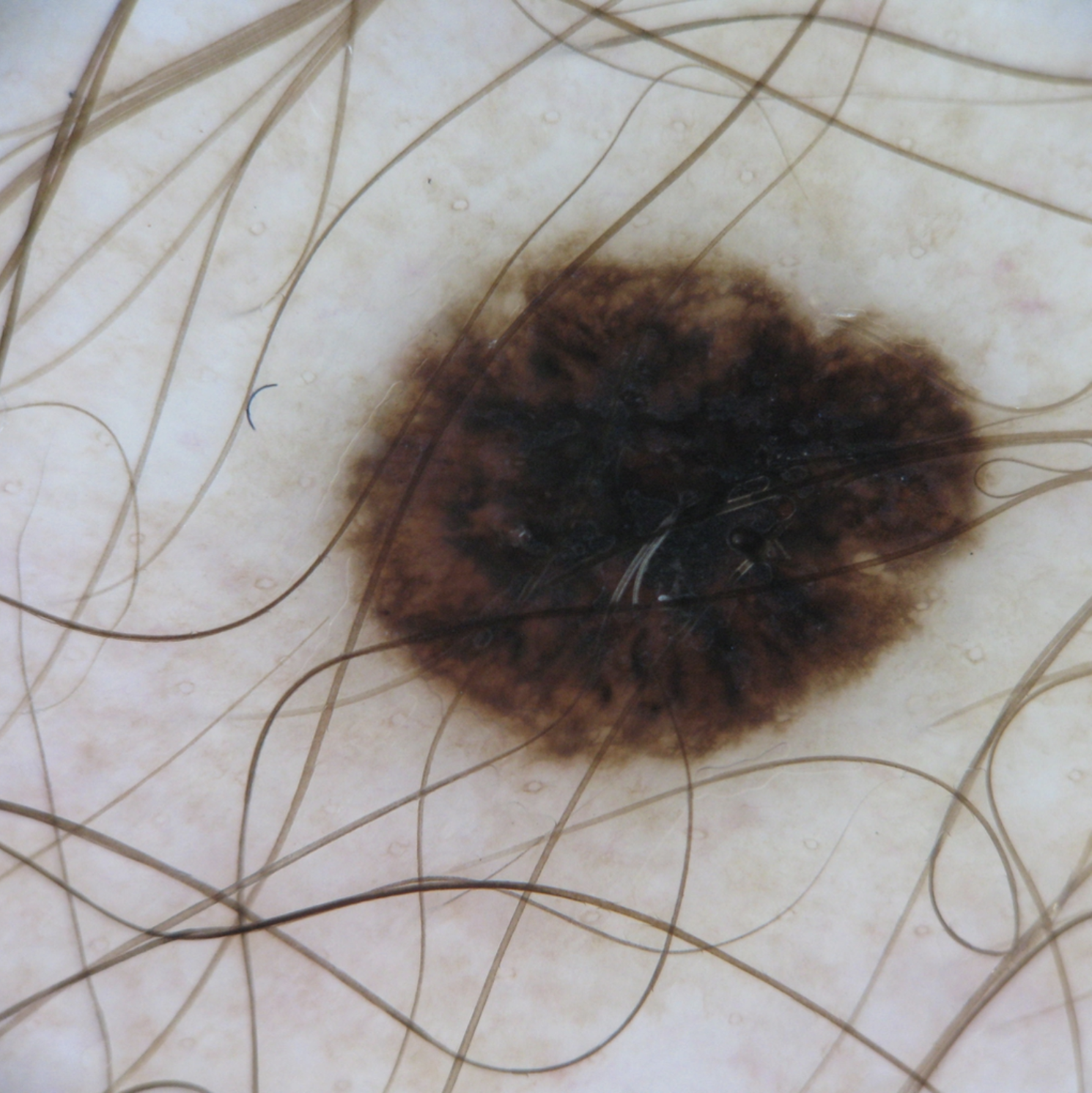}}
    & ISIC2019~\cite{isic2019}      & 36.14 & 28.17 & 33.31 & 34.05 & \bestfig{36.94} & 29.66 & MK \\
    \raisebox{-0.35em}{\includegraphics[height=0.58cm]{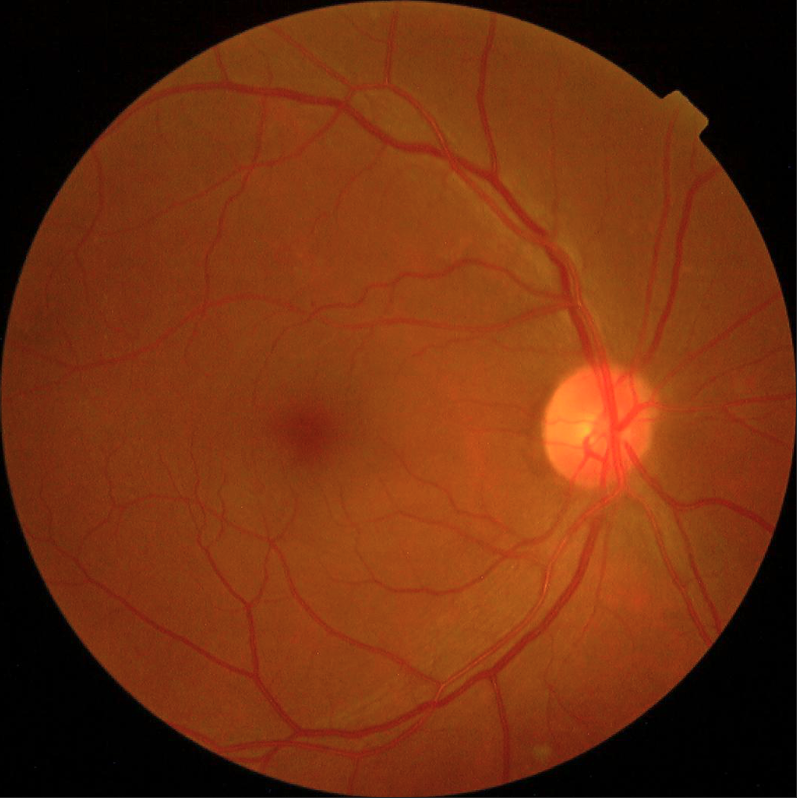}}
    & APTOS2019~\cite{aptos2019}     & 55.03 & 39.65 & 53.84 & 51.83 & 52.06 & \bestfig{57.74} & LBP \\
    \bottomrule
\end{tabular}}
\vspace{6pt}

\textbf{(b) Reasoning comparison on a BreakHis image}\par\vspace{4pt}
\noindent
\begin{minipage}[t]{0.135\linewidth}
    \centering
    \textbf{Input}\par\vspace{2pt}
    \begin{tcolorbox}[rbox, colback=gray!5, colframe=gray!50, valign=center]
    \textbf{Prompt:} ``Determine the best topological descriptor.''\\[4pt]
    \textbf{Input Image:}\\[2pt]
    \centering
    \includegraphics[width=\linewidth]{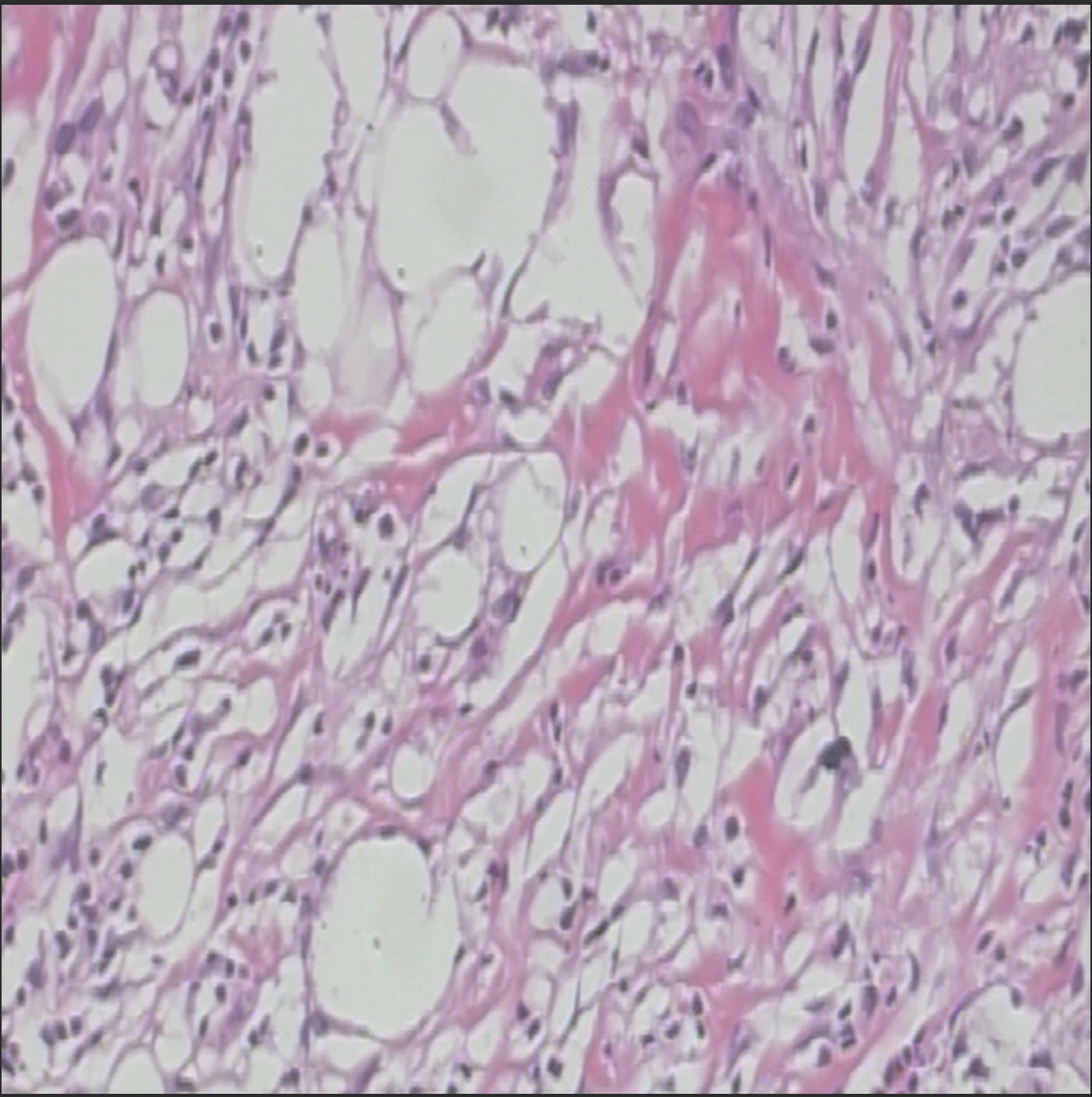}
    \end{tcolorbox}
\end{minipage}\hfill
\begin{minipage}[t]{0.185\linewidth}
    \centering
    \textbf{MedRAX~\cite{fallahpour2025medrax}}\par\vspace{2pt}
    \begin{tcolorbox}[rbox, colback=red!5, colframe=failRed]
    \textbf{Analysis:} ``$H_1 \rightarrow$ glandular lumens. Signal is \textit{noisy}.''\\[2pt]
    \textbf{Decision:} Better medical interpretation, but same noise-driven default.\\[2pt]
    \textbf{Result:} Pers.\ Silhouette~\textcolor{failRed}{\texttimes}
    \end{tcolorbox}
\end{minipage}\hfill
\begin{minipage}[t]{0.170\linewidth}
    \centering
    \textbf{Claude~\cite{anthropic2025claude}}\par\vspace{2pt}
    \begin{tcolorbox}[rbox, colback=red!5, colframe=failRed]
    \textbf{Analysis:} ``$H_1$ reflects loops from lumens and keratin whorls.''\\[2pt]
    \textbf{Decision:} Richest anatomy; no empirical evidence to verify.\\[2pt]
    \textbf{Result:} Pers.\ Landscape~\textcolor{failRed}{\texttimes}
    \end{tcolorbox}
\end{minipage}\hfill
\begin{minipage}[t]{0.185\linewidth}
    \centering
    \textbf{Gemini~\cite{google2025gemini}}\par\vspace{2pt}
    \begin{tcolorbox}[rbox, colback=red!5, colframe=failRed]
    \textbf{Analysis:} ``Low avg persistence $\rightarrow$ noise. $H_1$ max${=}$0.38 $\rightarrow$ stable loops.''\\[2pt]
    \textbf{Decision:} Sound TDA reasoning, but produces invalid tool parameters.\\[2pt]
    \textbf{Result:} Pers.\ Image~\textcolor{failRed}{\texttimes}
    \end{tcolorbox}
\end{minipage}\hfill
\begin{minipage}[t]{0.280\linewidth}
    \centering
    \textbf{TopoAgent}\par\vspace{2pt}
    \begin{tcolorbox}[rbox, colback=green!5, colframe=passGreen]
    \textbf{Perception:} \textit{glands/lumens}; balanced $H_0$/$H_1$.\\[1pt]
    \textbf{Reasoning:} Propose Pers.\ Image. {\color{skillGreen}Rankings}: Pers.\ Statistics is Tier~1 $\rightarrow$ switch. {\color{memGold}Memory}: confirmed (3 trials).\\[1pt]
    \textbf{Action:} Call Pers.\ Statistics.\\[1pt]
    \textbf{Reflection:} 0\% sparsity, high variance.\\[1pt]
    \textbf{Result:} Pers.\ Statistics~\textcolor{passGreen}{$\checkmark$}
    \end{tcolorbox}
\end{minipage}
    \end{minipage}
}

\caption{(a) Balanced accuracy (\%) of 6 descriptors (covering the top-3 per object type) on 5 datasets. PS\,=\,persistence statistics, PI\,=\,persistence image, TF\,=\,template function, ATOL\,=\,automatic topologically-oriented learning, MK\,=\,Minkowski functional, LBP\,=\,local binary pattern. The best descriptor (bold) differs across 5 datasets, confirming no single descriptor is best for all. \textbf{(b)} Reasoning comparison on a BreakHis image (Pers.\,=\,persistence). General-purpose LLMs with the same tools all select suboptimal descriptors; GPT-4o~\cite{openai2024gpt4o} (MedRAX's backbone) reaches similar result without medical insight. TopoAgent determines persistence statistics for glands/lumens.}
\label{fig:motivation}

\end{figure*}

Persistent homology (PH), a central tool in topological data analysis (TDA), captures geometric structural properties in images that conventional pixel-level deep learning (DL) approaches often neglect~\cite{edelsbrunner2002topological,carlsson2009topology,meng2026topocl,liang2025cell,adame2025topo}. In medical imaging, this topological perspective has proven valuable for grading cancer~\cite{lawson2019prostate,wang2023ccf}, predicting breast cancer survival~\cite{singhal2026topology}, analyzing retinal vasculature~\cite{ahmed2025topo}, and classifying tissue morphology~\cite{gu2025topoimages,gu2026integrating}, with growing adoption in other visual domains such as shape analysis and materials science~\cite{su2025topological}. To make these topological insights actionable for downstream tasks, researchers rely on \textit{topological descriptors}: methods that convert persistence diagrams (PDs) or raw images into fixed-length topological feature vectors~\cite{ali2023survey,su2025topological}. A rich family of such descriptors is known, including persistence images~\cite{adams2017persistence}, persistence 
landscape~\cite{bubenik2015statistical}, Euler characteristic transform~\cite{turner2014persistent}, and Minkowski functional~\cite{mecke2000additivity}, among others (see Supplementary for the complete descriptor taxonomy). However, each descriptor encodes distinct geometric characteristics and entails specific mathematical properties, hyperparameters, and failure modes. No single descriptor is universally effective across all datasets~\cite{ali2023survey} (e.g., see Fig.~\ref{fig:motivation}(a)). Thus, determining effective descriptors for a given image dataset is an overwhelming challenge that demands a great deal of expertise and effort.

This determination is challenging because it depends on the interaction between dataset-specific morphological characteristics (e.g., connected components, loops, cavities) and each descriptor's mathematical properties. In practice, users must manually analyze dataset properties, choose among candidate descriptors, tune hyperparameters, and produce topological feature vectors for downstream tasks. With a large descriptor library and continuous parameter spaces, exhaustive evaluation is prohibitive at the per-image level, making this trial-and-error process time-consuming, expertise-dependent, and difficult to standardize.

Large language model (LLM)-based agentic frameworks offer a ``natural'' solution. Modern agents integrate reasoning, tool use, memory, and iterative refinement~\cite{yao2022react,shinn2023reflexion,zhu2025fmri2ges,zhu2026anatomy}, and medical AI agents have begun orchestrating domain-specific tools for clinical analysis~\cite{fallahpour2025medrax,li2024mmedagent,wang2025medagent,xu2025comprehensive, xiao2026not}. Since descriptor determination requires both high-level reasoning about data morphology and low-level computation (e.g., PH and descriptor extraction), it aligns naturally with this paradigm. However, no existing agentic framework automatically computes PH, reasons about topological structures, and produces topological feature vectors for downstream tasks. General-purpose LLMs can describe topology in natural language but cannot compute it without external tool integration~\cite{liu2024chatgpt}, and even with tools, they lack the empirical grounding to verify their determinations (e.g., see Fig.~\ref{fig:motivation}(b)).

Automating descriptor determination also requires systematic evaluation, but prior studies typically assess a small number of descriptors on limited datasets with varying preprocessing, classifiers, and evaluation metrics~\cite{ali2023survey}, making cross-study comparisons unreliable. No standardized benchmark spans diverse medical image morphologies with consistent criteria, leaving the community without clear guidance on when specific topological descriptors excel or fail.

To address these gaps, we propose \textbf{TopoAgent} and \textbf{TopoBenchmark}. TopoAgent is an LLM-based agentic framework that automates topology learning for medical images by determining the most suitable topological descriptors for an input dataset. Given a raw image and a task prompt, TopoAgent operates through a Perception--Reasoning--Action--Reflection (PRAR) loop without task-specific training. Perception characterizes the image's topological and visual properties using perception tools. Reasoning first formulates a descriptor proposal by matching the observed characteristics based on descriptor properties, and then integrates empirical evidence from a distilled skill set and dual memory to determine the final descriptor and parameters. 
Action executes the determined descriptor using descriptor tools to produce a topological feature vector. Reflection validates the output quality and, on failure, records a diagnosis in long-term memory and triggers retry, enabling the agent to self-correct rather than committing to a single-shot decision. TopoBenchmark is a standardized evaluation benchmark comprising 26 datasets that broadly cover publicly available 2D medical imaging scenes, organized into five object types that span the major morphological categories in medical images: \textit{cells}, \textit{glands and lumens}, \textit{organ shapes}, \textit{vessel trees}, and \textit{surface lesions}. It provides an empirical basis from which the skill set is distilled and a frozen testbed of 113,182 samples with convergence-based per-dataset sample sizing for consistent evaluation.

Our contributions are threefold. (1)~We introduce TopoAgent, the first agentic framework that bridges LLM-based reasoning and topological data analysis, enabling automated topology learning from raw medical images with full reasoning traces. (2)~We construct TopoBenchmark, the first standardized benchmark for evaluating topological descriptors across diverse medical image morphologies with consistent criteria. (3)~We demonstrate that adaptive, per-image descriptor determination outperforms the strongest baseline by 9.32\% and general-purpose LLMs equipped with the same tools by over 21\% in average balanced accuracy. Our code is publicly available at \url{https://github.com/gm3g11/TopoAgent}.

\section{Background and Related Work}
\label{sec:background}

\subsection{Topological Descriptors}
\label{sec:bg_tda}

PH tracks the appearance and disappearance of topological features across a filtration of the input image $I$. For 2D medical images, we apply a sublevel-set filtration on pixel intensities via cubical complexes, yielding a PD $\mathrm{Dgm}_k(I)$: a multiset of birth-death pairs $\{(b_i, d_i)\}$ recording when each $k$-dimensional feature appears ($b_i$) and vanishes ($d_i$). Here, $k=0$ captures connected components and $k=1$ captures loops. The persistence $d_i - b_i$ measures each feature's significance across scales~\cite{edelsbrunner2002topological,carlsson2009topology}. To make PDs compatible with machine learning, topological descriptors convert them into fixed-length vectors. These fall into two families: PH-based descriptors $\varphi_d(\mathrm{Dgm}(I);\,\theta) \to \mathbb{R}^{n_d}$ that operate on PDs (e.g., persistence images~\cite{adams2017persistence}, persistence landscape~\cite{bubenik2015statistical}, Betti curve~\cite{umeda2017time}), and image-based descriptors $\varphi_d(I;\,\theta) \to \mathbb{R}^{n_d}$ that extract topological or geometric features directly from the image, bypassing PH entirely (e.g., Euler characteristic transform~\cite{turner2014persistent}, Minkowski functional~\cite{mecke2000additivity}). Each descriptor encodes different structural properties with distinct mathematical properties, and no single descriptor is universally effective across all datasets~\cite{ali2023survey}. We refer readers to~\cite{ali2023survey,su2025topological} for comprehensive surveys of topological descriptors. From these surveys and
literature~\cite{meng2025efficient, chambers2025stable, zia2024topological,  ahmed2025topological}, we identify 15 descriptors that are
widely adopted, mathematically well-founded, and applicable
to 2D medical images, forming the descriptor library
$\mathcal{D}$ used throughout this work. Recent work has integrated topological features into DL pipelines for image classification through differentiable topological layers~\cite{carriere2020perslay,kim2020pllay}, notably PHG-Net~\cite{peng2024phg}, which fuses learned persistence features with CNN and Transformer backbones. These approaches learn a fixed vectorization end-to-end rather than determining which descriptor suits a given input.

\subsection{LLM-based Agentic Frameworks}
\label{sec:bg_agents}
LLM-based agents extend LLMs beyond text generation by equipping them with tools, memory, and structured reasoning loops. ReAct~\cite{yao2022react} introduced the paradigm of interleaving chain-of-thought reasoning with tool calls, enabling models to ground their reasoning in external observations. Reflexion~\cite{shinn2023reflexion} added self-reflection, where agents learn from past failures stored in verbal memory to improve subsequent attempts. Voyager~\cite{wang2023voyager} demonstrated open-ended skill acquisition through a growing library of reusable behaviors. These ideas have been synthesized into the PRAR architecture pattern 
\cite{zhu2026ants,xi2025rise,khamis2025agentic,meng2025psychology,zhu2024advancing,yu2026i2eimagepixelsactionable}, which our TopoAgent adopts as the backbone. More broadly, skills in agentic AI are reusable knowledge modules that encode domain expertise to augment an agent's reasoning~\cite{wang2024survey,xi2025rise}; they range from executable code snippets to structured decision rules. In medical AI, agentic systems have begun orchestrating domain-specific tools: MedRAX~\cite{fallahpour2025medrax} coordinates radiological analysis tools for chest X-ray reasoning, MMedAgent~\cite{li2024mmedagent} routes queries across multi-modal medical tools, and MedAgent-Pro~\cite{wang2025medagent} chains diagnostic workflows with tool-augmented reasoning~\cite{zhu2025pathology,zhu2026medeyes,lin2026models,zhu2026medsynapse}. However, no such frameworks compute PH, reason about topological structures, and produce topological feature vectors. An alternative to agentic determination is automated machine learning (AutoML)~\cite{feurer2015efficient,hutter2019automated}, which searches parameter spaces via black-box optimization. However, AutoML treats descriptor choice as a hyperparameter to tune rather than a reasoning problem that requires understanding of the interaction between data topology and descriptor properties, and thus cannot generalize from prior experience to unseen inputs without rerunning the search.

\section{Method}\label{sec:method}

\begin{figure*}[t]
\centering
\includegraphics[width=1\textwidth, trim=0 222 182 0, clip]{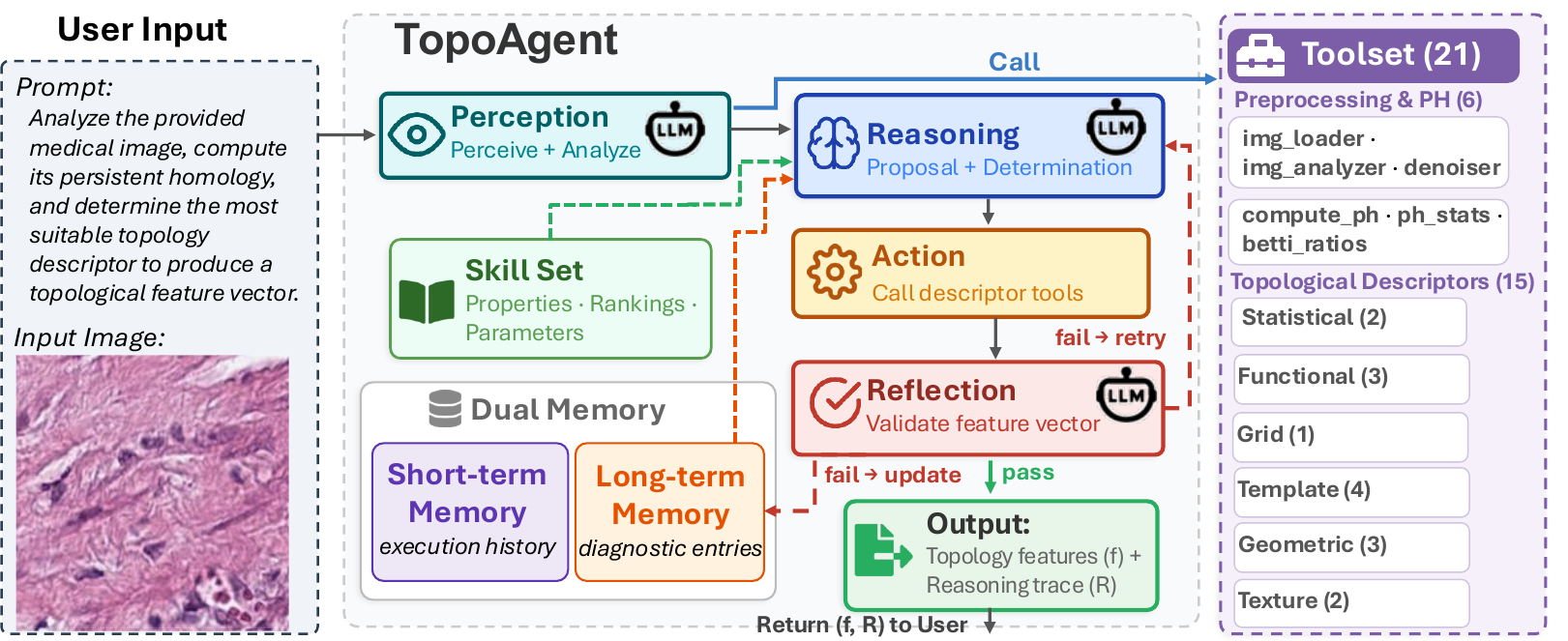}
\caption{An overview of the TopoAgent framework with four phases: \textbf{Perception} identifies the object type and analyzes persistent homology statistics; \textbf{Reasoning} proposes and 
determines the descriptor and parameters; \textbf{Action} calls the descriptor tools; \textbf{Reflection} validates the feature vector. The skill set $\mathcal{S}$ feeds descriptor properties and tiered rankings with parameters to Reasoning (dashed green). Short-term memory $M_s$ tracks execution history within a run; long-term memory $M_l$ supplies diagnostic entries to Reasoning (dashed orange) and is updated on failure.}
\label{fig:framework}
\vspace{-0.3cm}
\end{figure*}

\subsection{Framework Overview}\label{sec:overview}
Given a medical image $I$ and a task prompt, TopoAgent determines the most suitable topological descriptor $d^* \in \mathcal{D}$ from a library of $|\mathcal{D}|=15$ descriptors, configures its parameters $\theta^*$, and produces a topological feature vector $\mathbf{f} \in \mathbb{R}^{n_d}$ along with a reasoning trace $\mathcal{R}$ documenting all decisions. 
It uses the PRAR architecture~\cite{wang2024survey,xi2025rise,khamis2025agentic}, extended with tools, a distilled skill set $\mathcal{S}$, and dual memory ($M_s$, $M_l$). The skill set $\mathcal{S}$ comprises three components: descriptor properties $\mathcal{S}_{\mathrm{prop}}$ encoding qualitative knowledge, empirical evidence $\mathcal{S}_{\mathrm{rank}}$ providing tiered rankings and reasoning chains, and validated parameters $\mathcal{S}_{\mathrm{param}}$ specifying optimal configurations per object type. Short-term memory $M_s$ records tool invocations within each run; long-term memory $M_l$ keeps diagnostic entries across runs.
Fig.~\ref{fig:framework} shows the 
workflow. 
We present the four-phase pipeline in \S\ref{sec:pipeline} and 
the supporting components in \S\ref{sec:components}.

\subsection{TopoAgent Pipeline}\label{sec:pipeline}

{\bf Perception.}
Before reasoning about descriptor suitability, the agent must first understand the image: its imaging modality, tissue or anatomical structures, and topological characteristics. Perception invokes six perception tools deterministically, producing three outputs. (a) The PH profile $h(I)$ summarizes the topological characteristics: birth-death pair counts per homology dimension, average persistence, Betti ratio $\beta_1/\beta_0$ (the ratio of loop count to component count), and per-dimension persistence distributions. (b) Visual statistics $v(I)$ capture image-level properties such as signal-to-noise ratio, contrast, and edge density. (c) The object type $o \in \mathcal{O} = \{\textit{cells}, \textit{glands/lumens}, \textit{organ shapes}, \textit{vessel trees}, \textit{surface lesions}\}$ is identified from the image content. Deterministic tool execution ensures reproducible PH computation. The LLM's role in Perception is interpretive: it jointly reasons on the raw image and the computed PH to identify $o$. The object type $o$ is critical because it indexes the skill set: empirical evidence $\mathcal{S}_{\mathrm{rank}}$ and validated parameters $\mathcal{S}_{\mathrm{param}}$ are all organized per object type.

Both topological and visual 
information are necessary for reliable object-type identification. The PH profile quantifies topological characteristics that directly inform descriptor choice, such as the relative prevalence of connected components versus loops and their persistence distributions. However, the PH profile alone cannot reliably distinguish all object types. For example, both dense cell populations and glandular tissues with fragmented lumen boundaries can give somewhat similar $H_0$-dominated PH profiles, but they require fundamentally different descriptors due to distinct underlying morphology. Visual grounding resolves such ambiguities: jointly observing the image and its PH substantially outperforms both PH-profile-only and vision-only identification (Table~\ref{tab:ablation}), confirming that the two information sources provide complementary evidence.

\noindent
{\bf Reasoning.}
Reasoning is the central phase of the pipeline, responsible for both proposing and determining the descriptor configuration. It addresses the core challenge of the framework:
determining an effective topological descriptor requires understanding of the interaction between the image's topological characteristics and each descriptor's mathematical properties. To handle this, Reasoning 
takes two steps with asymmetric information access.

In the \textit{proposal step}, the agent receives the Perception outputs $(h(I), v(I), o)$ together with $\mathcal{S}_{\mathrm{prop}}$, which encodes the mathematical definition, strengths, weaknesses, and intended use of each descriptor. Crucially, $\mathcal{S}_{\mathrm{rank}}$ is provided in stripped form: the agent sees reasoning patterns that describe how PH characteristics relate to descriptor suitability, but not the descriptor recommendations or tiered rankings themselves. The agent reasons about which descriptor's mathematical properties best match the observed topological characteristics and proposes a candidate $(d_p, \theta_p)$ with preliminary parameters. In the \textit{determination step}, the agent integrates the full $\mathcal{S}_{\mathrm{rank}}$ (including tiered rankings, descriptor recommendations, and threshold-based PH signal definitions) and long-term memory $M_l$ (outcomes from prior runs on the current dataset). Weighing the proposal against this empirical evidence, the agent arrives at one of three decisions: it retains $(d_p, \theta_p)$ when the PH profile provides strong evidence despite a low ranking, defers to $\mathcal{S}_{\mathrm{rank}}$ when the proposal is weakly supported, or adopts a correction from $M_l$ that overrides both.

This two-step design addresses anchoring bias, the tendency of LLMs to fixate on the first salient information that they receive rather than reasoning independently. When the full $\mathcal{S}_{\mathrm{rank}}$ is visible during proposal formation, the LLM determines the top-ranked descriptor in 92.1\% of 520 held-out test images (20 per dataset in 26 datasets), even when explicitly prompted to ``refer to but not rely on'' the rankings. This reduces the agent into a lookup table. Providing only stripped reasoning patterns during the proposal step reduces the top-ranked agreement rate to 61.3\%: of these, 38.8\% are cases where the agent independently arrived at the same determination, while 22.5\% are cases where $\mathcal{S}_{\mathrm{rank}}$ corrected the proposal during determination (more details in the Supplementary).

\noindent
{\bf Action.}
Action executes the determined descriptor $d^*$ with parameters $\theta^*$ by invoking the corresponding descriptor tools, producing a topological feature vector $\mathbf{f} \in \mathbb{R}^{n_d}$. The parameters are drawn from the ranges validated during skill set construction for the identified object type $o$ rather than generated by the LLM, ensuring reproducible and correctly configured tool calls.

\noindent
{\bf Reflection.}
A single pass through Reasoning and Action does not guarantee a suitable output: the determined descriptor may produce an unsuitable feature vector due to parameter misconfiguration (e.g., overly fine resolution causing excessive sparsity), a mismatch with the image's topological characteristics, or edge-case PH characteristics not anticipated by~$\mathcal{S}$. Reflection addresses this by having the LLM assess the quality of~$\mathbf{f}$ based on summary statistics (sparsity, variance, kurtosis, skewness, dynamic range, and informative-feature ratio) together with descriptor-specific reference ranges from~$\mathcal{S}$ and diagnostic entries from~$M_l$. Crucially, reference ranges are provided as context, not as hard thresholds: what constitutes acceptable quality varies across descriptors and tissue types. For instance, high sparsity in a persistence image may be appropriate for sparse PH but indicates failure on dense glandular tissue; LBP histograms naturally exhibit high kurtosis that fixed thresholds would flag as anomalous. If the LLM diagnoses a quality failure, it returns a structured correction (e.g., ``switch to a lower-dimensional descriptor''), recorded in~$M_l$ along with the failed descriptor and diagnosed cause, then retries from the determination step of Reasoning. A maximum of two retries is permitted within~$t_{\max}$; if no passing result is obtained, the agent falls back to the top-ranked descriptor for~$o$ from~$\mathcal{S}_{\mathrm{rank}}$. Reflection validates feature vector quality, not downstream classification accuracy, since class labels are unavailable at inference time. On success, TopoAgent returns~$\mathbf{f}$ together with a reasoning trace $\mathcal{R} = \{h(I), v(I), o, d_p, d^*, \theta^*, q\}$, where~$q$ denotes the quality diagnosis.

\subsection{TopoAgent Components}\label{sec:components}

{\bf Toolset.} The agent accesses 21 domain-specific tools organized into two groups. 6~\textit{perception tools} support the Perception phase: image loading, quality analysis, denoising, PH computation via GUDHI~\cite{gudhi} with cubical complex filtration, PH profile extraction, and Betti ratio computation. 15~\textit{descriptor tools} support the Action phase: 10 PH-derived descriptors that vectorize persistence diagrams (e.g., persistence image~\cite{adams2017persistence}, persistence landscape~\cite{bubenik2015statistical}, ATOL~\cite{royer2021atol}) and 5 image-based descriptors that capture geometric properties without explicit PH computation (e.g., Minkowski functional~\cite{mecke2000additivity}, Euler characteristic transform~\cite{turner2014persistent}, LBP texture~\cite{ojala2002multiresolution}); see Supplementary for the complete taxonomy. Each tool exposes configurable parameters whose optimal values per object type are determined during skill set construction. Implementing descriptors as pre-built, parameterized tools ensures that the LLM reasons about descriptor suitability rather than generating computation code.

\noindent
{\bf Skill Set.}
To determine suitable descriptors, TopoAgent needs domain expertise that LLMs lack: understanding of each descriptor's mathematical properties, empirical evidence of descriptor effectiveness per object type, and appropriate parameter configurations. We encode this expertise in a static skill set $\mathcal{S}$, distilled offline from systematic evaluation on TopoBenchmark (\S\ref{sec:topobenchmark}): descriptor properties $\mathcal{S}_{\mathrm{prop}}$, empirical evidence $\mathcal{S}_{\mathrm{rank}}$, and validated parameters $\mathcal{S}_{\mathrm{param}}$.

As shown in Fig.~\ref{fig:skillset}, $\mathcal{S}_{\mathrm{prop}}$ encodes qualitative knowledge for the proposal step: the mathematical definition, strengths, weaknesses, and intended use of each descriptor, along with parameter reasoning heuristics that guide the LLM toward appropriate configurations given the observed PH profile. $\mathcal{S}_{\mathrm{rank}}$ provides empirical evidence for the determination step: per-object-type descriptor rankings presented as tiers rather than exact accuracy values, reasoning chains that map PH profile patterns to descriptor recommendations, and threshold-based PH signal definitions. As detailed in \S\ref{sec:pipeline}, the proposal step receives reasoning patterns without descriptor recommendations, while the determination step receives the full evidence including tiered rankings. $\mathcal{S}_{\mathrm{param}}$ specifies optimal dimensionality controls and additional tunable parameters for all 75 descriptor--object-type combinations, serving as reference configurations for the LLM during Reasoning and as deterministic fallback for Action (see Supplementary for full details).

$\mathcal{S}$ is constructed using classification accuracy as the evaluation metric, a standard protocol for topological descriptors~\cite{ali2023survey}. We evaluate on 26 TopoBenchmark datasets (\S\ref{sec:topobenchmark}) with 6 classifiers drawn from the top of the TabArena leaderboard~\cite{erickson2025tabarena} (TabPFN~\cite{hollmann2025tabpfn}, XGBoost~\cite{chen2016xgboost}, CatBoost~\cite{prokhorenkova2018catboost}, Random Forest \cite{breiman2001random}, TabM~\cite{gorishniy2025tabm}, RealMLP~\cite{holzmuller2024better}) and 5-fold cross-validation. The construction proceeds in three stages.  (A) Grid search identifies optimal parameter and dimensionality configurations for each descriptor on each dataset, yielding $\mathcal{S}_{\mathrm{param}}$. (B) Using these optimized configurations, we evaluate all 15 descriptors across 26 datasets and 6 classifiers ($26 \times 15 \times 6 = 2{,}340$ balanced accuracy values), and derive per-object-type tiered rankings from each descriptor's best-performing classifier per dataset, averaged across datasets of the same object type, yielding $\mathcal{S}_{\mathrm{rank}}$. (C) Five domain experts analyzed the quantitative results and formulated the qualitative components of both $\mathcal{S}_{\mathrm{prop}}$ and $\mathcal{S}_{\mathrm{rank}}$: descriptor properties, parameter reasoning heuristics, reasoning chains, and PH signal definitions, all grounded in TDA principles.
To prevent information leakage and improve generalization, all components of $\mathcal{S}$ are organized at the object-type level rather than encoding dataset-specific information. The agent never sees dataset identities, class labels, or per-dataset accuracy values at inference time. For example, $\mathcal{S}_{\mathrm{rank}}$ records the persistence landscape ranks in Tier~1 for \textit{glands/lumens}, not that it attains a specific accuracy on any dataset. While $\mathcal{S}$ shares similarities with 
retrieval-augmented generation (RAG), two key differences 
distinguish it: asymmetric information access during 
proposal prevents anchoring (\S\ref{sec:pipeline}), 
and Reflection validates outputs against 
quality criteria, which standard RAG pipelines lack.

\begin{figure*}[t]
\centering
\resizebox{1\textwidth}{!}{%
\begin{tikzpicture}[
    every node/.style={font=\small},
    header/.style={
        fill=headerbg,
        text=white,
        font=\small\bfseries,
        minimum height=0.55cm,
        rounded corners=3pt,
        inner sep=3pt,
        anchor=north,
    },
    exbox/.style={
        rounded corners=2pt,
        inner sep=4pt,
        font=\footnotesize,
        align=left,
        anchor=north,
    },
]
\node[header, minimum width=4.6cm] (h1) at (0, 0)
    {$\mathcal{S}_{\mathrm{prop}}$: Descriptor Properties};
\node[below=3pt of h1, exbox, fill=propbg, draw=propborder,
      line width=0.4pt, text width=4.2cm] (ex1) {%
\textbf{Example: Persistence Image}\\[2pt]
\textbf{Best when:} Both $H_0$ and $H_1$ content\\[1pt]
\textbf{Weakness:} Wastes dimensions on empty $H_1$; $\sigma$ tuning critical\\[3pt]
\rule{\linewidth}{0.3pt}\\[2pt]
\textbf{Reasoning hint:}
\textit{``If avg persistence $<$ 0.01, use $\sigma$\,=\,0.5--0.6 to smooth short-lived noisy features.''}
};

\node[header, minimum width=6.0cm] (h2) at (5.4, 0)
    {$\mathcal{S}_{\mathrm{rank}}$: Empirical Evidence};
\node[below=3pt of h2, exbox, fill=rankbg, draw=rankborder,
      line width=0.4pt, text width=5.6cm] (ex2) {%
\textbf{Tiered rankings} (\textit{discrete\_cells}):\\[1pt]
\begin{tabular}{@{}l@{\ }l@{}}
Tier 1: & MinkowskiFn, TemplateFn\\
Tier 2: & EulerCurve, BettiCurves\\
Tier 3: & Silhouettes, PersEntropy\\
\end{tabular}\\[3pt]
\rule{\linewidth}{0.3pt}\\[2pt]
\textbf{Reasoning chain} ($H_0$-dominant):\\
\textit{``Discrete objects create rich $H_0$; signal is in $H_0$ birth/death distribution.''}\\[3pt]
\rule{\linewidth}{0.3pt}\\[2pt]
\textbf{PH signal rule:}
\textit{``$\beta_1/\beta_0 > 1.8 \wedge H_1$\,cnt $> 500$}
$\to$ \texttt{pers\_landscape}.\textit{''}
};

\node[header, minimum width=4.4cm] (h3) at (11.0, 0)
    {$\mathcal{S}_{\mathrm{param}}$: Validated Parameters};
\node[below=3pt of h3, exbox, fill=parambg, draw=paramborder,
      line width=0.4pt, text width=4.0cm] (ex3) {%
\textbf{Example: Persistence Image}\\[2pt]
\textit{vessel\_trees:}\\
\quad res=14, dim=392D\\
\quad $\sigma$=0.05, weight=squared\\[2pt]
\textit{glands\_lumens:}\\
\quad res=26, dim=1352D\\
\quad $\sigma$=0.6, weight=linear\\[3pt]
\rule{\linewidth}{0.3pt}\\[2pt]
\textit{$\sigma$ varies from 0.05 (thin vessels) to 0.6 (dense glandular tissue)}
};

\draw[decorate, decoration={brace, amplitude=6pt, raise=4pt},
      darktext, line width=0.7pt]
    ([yshift=3pt]h1.north west) -- ([yshift=3pt]h3.north east)
    node[midway, above=10pt, font=\small\bfseries,
         text=darktext] {Skill Set $\mathcal{S}$};
\end{tikzpicture}%
}
\caption{Examples of the skill set $\mathcal{S}$
using persistence image. $\mathcal{S}_{\mathrm{prop}}$ encodes qualitative descriptor knowledge, $\mathcal{S}_{\mathrm{rank}}$ provides tiered rankings and reasoning chains, and $\mathcal{S}_{\mathrm{param}}$ specifies validated parameters per object type.}
\label{fig:skillset}
\vspace{-0.3cm}
\end{figure*}

\noindent
{\bf Dual Memory.}
The static skill set cannot anticipate every possible scenario: unseen imaging modalities, atypical PH profiles, or edge cases may require runtime adaptation. TopoAgent addresses this through dual memory, inspired by the verbal memory mechanism in Reflexion~\cite{shinn2023reflexion}. Short-term memory$M_s$ records tool invocations within a single run, implemented as the accumulated history within the LLM's context window, preventing repeated failures on the same image. Long-term memory $M_l$ accumulates structured diagnostic entries across runs within a dataset: each entry records the failed descriptor, diagnosed cause, and successful correction, enabling the agent to learn from earlier mistakes on the same dataset. $M_l$ is reset between datasets to prevent cross-dataset leakage but accumulates within each dataset, enabling within-dataset adaptation.

Algorithm~\ref{alg:topoagent} summarizes the complete TopoAgent workflow, showing how the pipeline (\S\ref{sec:pipeline}) and components (\S\ref{sec:components}) interact.

\begin{algorithm}[t]
\caption{\textsc{TopoAgent}}\label{alg:topoagent}
\resizebox{1\linewidth}{!}{%
\begin{minipage}{\linewidth}
\begin{algorithmic}[1]
\Require Medical image $I$, task prompt, time limit $t_{\max}$
\Ensure Topological feature vector $\mathbf{f} \in \mathbb{R}^{n_d}$, reasoning trace $\mathcal{R}$
\State $\mathcal{S} \gets \Call{LoadSkillSet}{\,}$; \; $\mathit{Tools} \gets \{\text{perception tools (6)},\; \text{descriptor tools (15)}\}$
\State $M_s \gets [\,]$; \; $M_l \gets \Call{LoadLongTermMemory}{\,}$
\Statex \textbf{--- Perception ---}
\State $(h(I),\; v(I)) \gets \Call{PerceptionTools}{I}$; \; $M_s \gets M_s \cup \{h(I),\; v(I)\}$
\State $o \gets \Call{LLM.Perceive}{I,\; M_s}$ \Comment{Object type $o \in \mathcal{O}$ via vision-enabled LLM}
\Statex \textbf{--- Reasoning (Proposal) ---}
\State $(d_p,\; \theta_p) \gets \Call{LLM.Propose}{h(I),\; v(I),\; o,\; \mathcal{S}_{\mathrm{prop}}}$ \Comment{Stripped $\mathcal{S}_{\mathrm{rank}}$ only}
\Statex \textbf{--- Reasoning (Determination) + Action + Reflection ---}
\While{$\Call{Elapsed}{\,} < t_{\max}$}
    \State $(d^*,\; \theta^*) \gets \Call{LLM.Determine}{d_p,\; \theta_p,\; \mathcal{S}_{\mathrm{rank}},\; M_l}$
    \State $\mathbf{f} \gets \Call{ExecuteDescriptorTool}{d^*,\; \theta^*}$ \Comment{Action: run descriptor tool}
    \State $q \gets \Call{LLM.Reflect}{\mathbf{f},\; d^*,\; \mathcal{S},\; M_l}$ \Comment{Quality diagnosis}
\If{$q.\mathit{pass}$} \Return $(\mathbf{f},\; \{h(I), v(I), o, d_p, d^*, \theta^*, q\})$
\EndIf
\State $M_l.\mathit{update}(d^*,\; q)$
\Comment{Record diagnosis + correction}
\EndWhile
\State $(\mathbf{f},\; \mathcal{R}) \gets \Call{Fallback}{\mathcal{S}_{\mathrm{rank}},\; o}$ \Comment{Top-ranked descriptor for $o$}
\State \Return $(\mathbf{f},\; \mathcal{R})$
\end{algorithmic}
\end{minipage}%
}

\end{algorithm}

\subsection{TopoBenchmark}\label{sec:topobenchmark}
TopoBenchmark serves two purposes: it provides an empirical basis from which the skill set $\mathcal{S}$ is distilled, and it defines a frozen testbed for evaluating the agent's descriptor determinations. As discussed in \S\ref{sec:intro}, existing studies lack consistent protocols for cross-descriptor comparison. TopoBenchmark addresses this gap with standardized evaluation across diverse morphologies.

We curate 26 publicly available 2D medical image classification datasets spanning 11 imaging modalities and 11 clinical domains (full details in the Supplementary), organized into five object types that cover the major morphological categories in medical imaging: \textit{cells} (6 datasets), \textit{glands/lumens} (6), \textit{organ shapes} (8), \textit{vessel trees} (3), and \textit{surface lesions} (3). These five categories exhibit distinct topological signatures that motivate adaptive descriptor determination: discrete cells have the most balanced $H_0$/$H_1$ ratio ($\beta_1/\beta_0 \approx 1.3$) with the lowest total feature count (${\sim}$2,090 persistence pairs per image), vessel trees show the strongest $H_1$ dominance ($\beta_1/\beta_0 \approx 1.9$) from branching vasculature, and glands/lumens give the highest total feature count (${\sim}$4,842) from complex tissue architecture. Organ shapes exhibit the highest within-type variance, from 602 total pairs (OrganAMNIST) to 7,803 (OCTMNIST). No single descriptor performs best across all five types (Table~\ref{tab:topo_results}), showing that descriptor determination must be adaptive.

The complete collection contains over 1.2 million images, making full evaluation practically prohibitive. Moreover, the dataset
sizes vary by three orders of magnitude (516 to 327,680), making a uniform sample size inappropriate: too small wastes discriminative
power on large datasets, and too large exceeds the smallest datasets entirely. We address this with convergence analysis: for each dataset, we evaluate the top-3 descriptors at increasing sample sizes $n \in \{50, 100, 200, \ldots, 10{,}000\}$ ($n_{\max}$ capped at the dataset size) using 5-fold cross-validation with TabPFN~\cite{hollmann2025tabpfn} across 3 seeds, and define $n^*$ as the smallest $n$ satisfying three criteria simultaneously: balanced accuracy within 1\% of the value
at $n_{\max}$, standard deviation across the seeds below 2\%, and descriptor ranking agreement (Spearman $\rho = 1.0$ among the
top-3 descriptors). For example, PCam (5,000 histopathology patches) requires the full $n^* = 5{,}000$ for accuracy convergence, while IDRiD (516 retinal images) converges at
$n^* = 500$. However, descriptor ranking stabilizes much earlier ($n = 100$ for PCam, and $n = 50$ for IDRiD), meaning the relative ordering of the descriptors converges well before the absolute accuracy plateaus. The resulting frozen benchmark contains 113,182 samples with per-dataset sizes of 500--7,500, pre-computed persistent homology caches, and fixed fold indices for reproducible evaluation (the construction pipeline in the Supplementary).

\section{Experiments}\label{sec:experiments}
\textbf{Setup.} We evaluate TopoAgent on all 26 TopoBenchmark datasets grouped by the five object types. For each image, the agent receives only the raw image and a task prompt; no dataset identity or object-type label is provided. TopoAgent is implemented using LangGraph~\cite{langgraph} with GPT-4o~\cite{openai2024gpt4o} as its backbone LLM. We compare with three categories of baselines: (1)~general-purpose LLMs (GPT-4o, Gemini 2.5 Pro~\cite{google2025gemini}, Claude Sonnet 4.6~\cite{anthropic2025claude}), each equipped with the same 21 topology tools in a ReAct-style loop but without the skill set $\mathcal{S}$, dual memory, and the structured PRAR pipeline; (2)~a medical imaging agent (MedRAX~\cite{fallahpour2025medrax}) that orchestrates radiological tools but lacks topological capabilities\footnote{EndoAgent~\cite{tang2025endoagent} was also considered but excluded as it was withdrawn by its authors.}; (3)~fixed-descriptor baselines: persistence 
image~\cite{adams2017persistence}, persistence statistics~\cite{ali2023survey}, and an object-type oracle that selects the top-ranked descriptor from $\mathcal{S}_{\mathrm{rank}}$ with validated parameters 
from $\mathcal{S}_{\mathrm{param}}$ per object type, without per-image LLM reasoning. All LLM calls use greedy decoding (temperature $0$) for reproducibility. All evaluations use 5-fold cross-validation on frozen splits 
across six classifiers evaluated during skill set construction. We report the best accuracy 
among them. TabPFN~\cite{hollmann2025tabpfn} ranks the highest 
on the majority of the datasets. Implementation details are in the Supplementary.

\begin{table}[t]
    \centering
    \caption{Balanced accuracy (\%) on TopoBenchmark across the five object types. Bold: the best; \underline{underline}: the second best.}
    \label{tab:topo_results}
    \setlength{\tabcolsep}{6pt}
    \begin{tabular}{@{}l ccccc c@{}}
        \toprule
        \multirow{2}{*}{\textbf{Method}} 
        & \multicolumn{5}{c}{\textbf{TopoBenchmark}} 
        & \multirow{2}{*}{\textbf{Avg.}} \\
        \cmidrule(lr){2-6}
        & \textbf{Cells} & \textbf{Glands} & \textbf{Organs} & \textbf{Lesions} & \textbf{Vessels} & \\
        \midrule
        GPT-4o~\cite{openai2024gpt4o}              & 66.46 & 44.33 & 52.32 & 34.59 & 33.53 & 46.25 \\
        Gemini 2.5 Pro~\cite{google2025gemini}       & 63.57 & 52.19 & 52.10 & 33.39 & 33.53 & 46.96 \\
        Claude Sonnet 4.6~\cite{anthropic2025claude}    & 60.58 & 51.15 & 54.47 & 33.13 & 33.38 & 46.54 \\
        MedRAX~\cite{fallahpour2025medrax}          & 66.17 & 62.99 & 52.66 & 38.71 & 32.15 & 50.54 \\
        \midrule
        Persistence image~\cite{adams2017persistence}            & 59.17 & 47.22 & 44.03 & 35.04 & 36.14 & 44.32 \\ 
        Persistence statistics~\cite{ali2023survey}       & 64.43 & 54.34 & 52.73 & 42.76 & 34.76 & 49.80 \\ 
        Object-type oracle            & \underline{73.21} & \underline{72.16} & \underline{58.46} & \underline{50.13} & \underline{40.50} & \underline{58.89} \\
        \midrule
        \textbf{TopoAgent (ours)} 
        & \textbf{79.34} & \textbf{81.63} & \textbf{68.99} & \textbf{59.81} & \textbf{51.30} & \textbf{68.21} \\
        \bottomrule
    \end{tabular}

\end{table}

\noindent\textbf{Main Results.} Table~\ref{tab:topo_results} summarizes the balanced accuracy across the five object types. TopoAgent achieves 68.21\% average balanced accuracy, outperforming the strongest baseline (object-type oracle, 58.89\%) by 9.32\%. Since TopoAgent and the GPT-4o baseline share the same underlying LLM and toolset, the 21.96\% gap (46.25\% vs.\ 68.21\%) highlights the combined contribution of the structured PRAR pipeline, skill set, and dual memory. The three general-purpose LLMs cluster within a narrow range (46.25--46.96\%), suggesting that without structured reasoning, 
LLM capability alone is insufficient. MedRAX~\cite{fallahpour2025medrax} improves over the general-purpose LLMs by ${\sim}$4\% with medical knowledge but lacks topological features and falls nearly 18\% below TopoAgent. TopoAgent significantly outperforms all baselines under the Wilcoxon signed-rank test (oracle: $p < 0.01$; all others: $p < 0.001$). Among the fixed-descriptor baselines, the object-type oracle (58.89\%) represents the ceiling of non-adaptive determination; the 9.32\% gap to TopoAgent demonstrates the value of per-image descriptor determination, with the largest gains on vessels (+10.80\%) and organs (+10.53\%), where descriptor choice is most sensitive to image-specific PH characteristics.

\begin{table}[bt]
    \centering
\caption{Ablation study. Top: component 
ablation decomposing 
Perception, Reasoning, skill set, and dual memory. Middle: design ablation. 
Bottom: backbone ablation.}
    \label{tab:ablation}
    \setlength{\tabcolsep}{5pt}
    \begin{tabular}{@{}l ccccc c@{}}
        \toprule
        \multirow{2}{*}{\textbf{Variant}} 
        & \multicolumn{5}{c}{\textbf{TopoBenchmark}} 
        & \multirow{2}{*}{\textbf{Avg.}} \\
        \cmidrule(lr){2-6}
        & \textbf{Cells} & \textbf{Glands} & \textbf{Organs} & \textbf{Lesions} & \textbf{Vessels} & \\
        \midrule
        w/o Perception                 & 76.67 & 78.62 & 63.19 & 44.90 & 42.74 & 61.22 \\
        w/o Reflection                 & 75.18 & 80.62 & 60.15 & 50.90 & 41.55 & 61.68 \\
        w/o Skill set $\mathcal{S}$    & 71.68 & 79.79 & 57.81 & 51.57 & 36.96 & 59.56 \\
        w/o Dual memory                & 79.22 & 81.31 & 65.58 & 45.05 & 43.34 & 62.90 \\
        \midrule
        w/o PH profile $h(I)$              & 77.59 & 81.53 & 66.49 & 48.38 & 46.61 & 64.12 \\
        w/o Visual statistics $v(I)$       & 77.03 & 80.96 & 68.13 & 50.26 & 48.39 & 64.95 \\
        w/o Proposal step                  & 73.69 & 73.21 & 58.04 & 52.11 & 41.75 & 59.76 \\
        w/o $\mathcal{S}_{\mathrm{prop}}$   & 76.83 & 80.08 & 63.93 & 56.93 & 42.77 & 64.11 \\
        w/o $\mathcal{S}_{\mathrm{rank}}$  & 72.52 & 79.82 & 59.26 & 53.85 & 39.19 & 60.93 \\
        w/o $\mathcal{S}_{\mathrm{param}}$ & 74.21 & 79.99 & 62.41 & 55.29 & 41.25 & 62.63 \\
        w/o Short-term $M_s$      & 79.25 & 81.42 & 67.81 & 55.35 & 49.82 & 66.73  \\ 
        w/o Long-term $M_l$               & 79.26 & 81.36 & 66.42 & 48.89 & 46.34 & 64.45 \\             
        \midrule
        TopoAgent (Claude 4.6)       & 78.15 & \textbf{82.17} & \textbf{70.42} & 57.29 & \textbf{53.82} & \textbf{68.37} \\
        TopoAgent (Gemini 2.5 Pro)   & 76.93 & 80.95 & 70.25 & 59.14 & 52.65 & 67.98 \\
        TopoAgent (Llama3 8B~\cite{grattafiori2024llama})        & 70.98 & 76.63 & 56.41 &  41.63     &    40.89   &  57.31     \\
        \midrule
        {TopoAgent (GPT-4o)}  & \textbf{79.34} & 81.63 & 68.99 & \textbf{59.81} & {51.30} & 68.21 \\
        \bottomrule
    \end{tabular}
    
\end{table}

\noindent\textbf{Ablation Study.} Table~\ref{tab:ablation} examines the contribution of each TopoAgent component. Reasoning and Action are not ablated as they are essential for producing any output. Among the four removable components, the skill set $\mathcal{S}$ is the most impactful (8.65\% drop); without it, the agent is only marginally better than object-type oracle (59.56\% vs.\ 58.89\%), confirming that $\mathcal{S}$ provides an essential knowledge base. Perception (6.99\% drop) is critical for lesions and vessels, where object-type identification drives descriptor choice. Reflection (6.53\% drop) matters the most for organs and lesions, which quite often require parameter correction. Dual memory shows the smallest component-level drop (5.31\%) but is critical for lesions (14.76\%) and vessels (7.96\%), whose PH profiles deviate the most from the skill set's general rankings. The design ablation further decomposes each component. Among the skill set components, $\mathcal{S}_{\mathrm{rank}}$ individually has the largest impact (7.28\% drop), followed by $\mathcal{S}_{\mathrm{param}}$ (5.58\%) and $\mathcal{S}_{\mathrm{prop}}$ (4.10\%), confirming that empirical rankings carry more weight than descriptive properties. Removing the Proposal step drops the performance to 59.76\%, 
very close to object-type oracle (58.89\%), validating the anti-anchoring design: without an independent proposal, the agent reduces to following $\mathcal{S}_{\mathrm{rank}}$ directly, losing the per-image adaptation ability that accounts for the 9.32\% gap. The backbone ablation shows a narrow spread across 
the models (67.98--68.37\%), reflecting the design intent: $\mathcal{S}$ encodes domain expertise that any frontier LLM can leverage. However, Llama3 8B~\cite{grattafiori2024llama} drops to 57.31\%, below object-type oracle, indicating that frontier-level reasoning is necessary to benefit from the pipeline. The agent's value lies in the PRAR pipeline that orchestrates tool use, skill retrieval, and quality validation, which zero-shot LLMs lack (trailing by over 21\%).

\begin{figure*}[t!]
\centering
\input{figures/case_study}
\caption{Two case studies on \textit{organ shapes} with opposite PH profiles. \textbf{Case~A:} sparse but meaningful PH (avg persistence{=}0.06) --- the agent retains a Tier~4 proposal with data-driven parameters. \textbf{Case~B:} noisy PH (avg persistence$<$0.01) --- the agent abandons PH-based descriptors entirely. Same object type, same ranking, opposite determinations driven by image-specific PH signals.}
\label{fig:case_study}
\vspace{-0.6cm}

\end{figure*}

\noindent\textbf{Case Study.} In Fig.~\ref{fig:case_study}, two \textit{organ shape} images  share the same object type and the same $\mathcal{S}_{\mathrm{rank}}$ recommendation (\texttt{minkowski\_functionals}, Tier~1), yet the agent arrives at different determinations. In Case~A (avg persistence = 0.06), the agent retains a Tier~4 proposal based on sparse but meaningful PH and memory confirmation. In Case~B (avg persistence $<$ 0.01), the agent abandons PH-based descriptors entirely in favor of texture, guided by $M_l$. A rule-based system would assign the same descriptor to both images based on object type alone. TopoAgent distinguishes them: Proposal reads the PH profile, Determination weighs it against rankings and memory, and Reflection verifies the outcome. 

\noindent\textbf{Downstream Integration.} Table~\ref{tab:downstream} evaluates whether TopoAgent's determinations improve DL pipelines by integrating the chosen descriptors into ResNet-152~\cite{he2016resnet} and SwinV2-B~\cite{liu2022swinv2} following the PHG-Net~\cite{peng2024phg} fusion protocol on ISIC 2018~\cite{zhuang2018isic}, Prostate Cancer~\cite{lawson2019prostate}, and CBIS-DDSM~\cite{lee2017cbisddsm}. The fixed persistence image consistently \textit{hurts} performance relative to the backbone alone, confirming that a mismatched descriptor can be worse than no topology at all. TopoAgent matches or exceeds PHG-Net on 4 of 6 backbone/dataset combinations and remains within one standard deviation on the other two, using pre-computed topological feature vectors that require no differentiable PH layer, while the determined Minkowski functionals provide interpretable geometric signatures (area, perimeter, Euler characteristic) that PHG-Net's learned features lack. Although TopoAgent determines Minkowski functionals for all three external datasets, the parameters differ per dataset, reflecting $\mathcal{S}_{\mathrm{prop}}$-guided reasoning on unseen data rather than $\mathcal{S}_{\mathrm{rank}}$ lookup, as no benchmark rankings exist for these datasets. All three datasets are external to TopoBenchmark and were not used during skill set construction, confirming that the agent generalizes beyond its training data.

\begin{table}[t]
    \centering
    \caption{Downstream integration with CNN and Transformer backbones. PI = fixed persistence image; PHG = PHG-Net's~\cite{peng2024phg} learnable topological features; TopoAgent = adaptively determined descriptor. Accuracy (\%) is reported as mean$\pm$std with 5 runs. TopoAgent determines Minkowski functionals for all three datasets.}
    \label{tab:downstream}
    \setlength{\tabcolsep}{5pt}
    \begin{tabular}{@{}l ccc@{}}
        \toprule
        \textbf{Method} & \textbf{ISIC} & \textbf{Prostate} & \textbf{CBIS-DDSM} \\
        \midrule
        ResNet-152~\cite{he2016resnet}             & 88.76$\pm$0.37 & 93.14$\pm$0.33 & 72.01$\pm$0.51 \\
        ResNet-152 + PI          & 87.91$\pm$0.11 & 93.05$\pm$0.12 & 70.45$\pm$0.18 \\
        ResNet-152 + PHG         & 89.98$\pm$0.21 & 93.98$\pm$0.31 & \best{74.51$\pm$0.29} \\
        ResNet-152 + TopoAgent   & \best{90.13$\pm$0.13} & \best{94.22$\pm$0.18} & 74.26$\pm$0.14 \\
        \midrule
        SwinV2-B~\cite{liu2022swinv2}              & 90.57$\pm$0.29             & 95.87$\pm$0.25             & 73.79$\pm$0.31              \\
        SwinV2-B + PI            & 90.45$\pm$0.13             & 95.86$\pm$0.16             & 73.13$\pm$0.18             \\
        SwinV2-B + PHG           & 91.88$\pm$0.18              & \best{98.55$\pm$0.31}              & 77.08$\pm$0.30             \\
        SwinV2-B + TopoAgent     & \best{92.01$\pm$0.23}              & 97.89$\pm$0.24              & \best{77.13$\pm$0.24}              \\
        \bottomrule
    \end{tabular}
    \vspace{-0.6cm}
\end{table}

\noindent\textbf{Efficiency.}
For 86.9\% of images, the PRAR pipeline completes in a single pass with 4 LLM calls and 29.0\,s wall-clock time. Reflection triggers a retry in the remaining 13.1\%, adding on average 1.8 calls and 22.1\,s. Overall, TopoAgent requires 4.2 LLM calls and 31.9\,s per image on average, making it practical for batch processing across large datasets.

\section{Conclusions}
\label{sec:conclusion}

We presented TopoAgent, the first LLM-based agentic framework for automated topology learning in medical imaging. By combining a Perception--Reasoning--Action--Reflection loop with a distilled skill set and dual memory, TopoAgent addresses a critical bottleneck: determining effective descriptors currently requires extensive domain expertise and trial-and-error experimentation. On TopoBenchmark (26 datasets, five object types), TopoAgent achieves 68.21\% average balanced accuracy, outperforming the strongest baseline by 9.32\% while providing interpretable topological feature vectors. Downstream integration with both CNN and Transformer backbones confirms that its pre-computed features match or exceed learnable topological representations without additional training.

As the skill set is distilled from 26 2D medical image datasets, it may not generalize to fundamentally different topological structures (e.g., 3D volumetric data or non-medical domains). The agent also inherits the latency and cost of LLM inference (${\sim}$32s per image). Future work includes extending TopoBenchmark to 3D images, incorporating the agent into end-to-end training pipelines, and exploring multi-descriptor fusion guided by the agent's reasoning.

\noindent\textbf{Acknowledgments.}
The authors gratefully acknowledge support from the National Science Foundation under Grants CCF-2444309 and CCF-2523787, and from the American Heart Association under Award 26AIREA1574568.

\bibliographystyle{splncs04}
\bibliography{ref}
\end{document}